\documentclass[nohyperref]{article}

\def\isaccepted{}

\usepackage{microtype}
\usepackage{graphicx}
\usepackage{subfigure}
\usepackage{booktabs} 

\usepackage{hyperref}

\usepackage{icml2022}

\usepackage{amsmath}
\usepackage{amssymb}
\usepackage{mathtools}
\usepackage{amsthm}
\usepackage{array}
\newcolumntype{H}{>{\setbox0=\hbox\bgroup}c<{\egroup}@{}}

\usepackage[capitalize,noabbrev]{cleveref}

\usepackage{graphicx}
\usepackage{caption}

\theoremstyle{plain}

\theoremstyle{definition}

\theoremstyle{remark}

\usepackage[textsize=tiny]{todonotes}

\newcommand{\database}{SIXPAQ}
\newcommand{\dbexpansion}{\textbf{S}ynthesized question \textbf{I}nterpretations to e\textbf{X}tend \textbf{P}robably \textbf{A}sked \textbf{Q}uestions}

\def\nodata{0}

\icmltitlerunning{}

\begin{document}

\twocolumn[
\icmltitle{Answering Ambiguous Questions with a Database of \\ Questions, Answers, and Revisions}



\icmlsetsymbol{equal}{*}

\begin{icmlauthorlist}
\icmlauthor{Haitian Sun}{goog,cmu}
\icmlauthor{William W. Cohen}{goog}
\icmlauthor{Ruslan Salakutinov}{cmu}
\end{icmlauthorlist}

\icmlaffiliation{goog}{Google DeepMind}
\icmlaffiliation{cmu}{Carnegie Mellon University}

\icmlcorrespondingauthor{Haitian Sun}{haitiansun@google.com}

\icmlkeywords{question answering, NLP}

\vskip 0.3in
]



\printAffiliationsAndNotice{}  

\begin{abstract}
    Many open-domain questions are under-specified and thus have multiple possible answers, each of which is correct under a different interpretation of the question.  Answering such ambiguous questions is challenging, as it requires retrieving and then reasoning about diverse information from multiple passages.  
    \ifx\nodata\undefined
    We present a new state-of-the-art for answering ambiguous questions that exploits a new resource: a large database of unambiguous questions generated from Wikipedia called \database{}\dbexpansion{}. \database{} is much larger than prior question databases, and has questions that are less ambiguous.
    \else
    We present a new state-of-the-art for answering ambiguous questions that exploits a database of unambiguous questions generated from Wikipedia.
    \fi
    On the challenging ASQA benchmark, which requires generating long-form answers that summarize the multiple answers to an ambiguous question, our method improves performance by 15\% (relative improvement) on recall measures and 10\% on measures which evaluate disambiguating questions from predicted outputs. 
    \ifx\nodata\undefined
    \database{} also dramatically improves answer recall relative to prior question databases (by around 30 points), and gives large improvements in diverse passage retrieval (by matching user questions $q$ to passages $p$ indirectly, via questions $q'$ generated from $p$).
    \else
    Retrieving from the database of generated questions also gives large improvements in diverse passage retrieval (by matching user questions $q$ to passages $p$ indirectly, via questions $q'$ generated from $p$).
    \fi
    
    \ifx\nodata\undefined
    We will release \database{} for future research.
    \fi
\end{abstract}

\section{Introduction}
Most previous research in open-domain QA focuses on finding the most probable answer to a question \citep{joshi2017triviaqa,kwiatkowski2019natural}. However,
many questions are ambiguous, and hence have multiple possible answers, each of which is correct under some interpretation of the question. For example, ``Where is the home stadium of the Michigan Wolverines?'' has different answers depending on whether one interprets the question as being about  ``football'' or ``basketball''.
Several datasets have been proposed recently that test models' ability to predict all possible answers \citep{zhang2021situatedqa,dhingra2022time,sun2022conditionalqa} and explain the differences between those answers \citep{min2020ambigqa,stelmakh2022asqa}.


The tasks of finding all possible answers, and finding all unambiguous interpretations of a question, have been shown to be very challenging for existing models, because these tasks require retrieving and aggregating diverse information \citep{min2020ambigqa,stelmakh2022asqa}.
The best-performing open-domain QA models typically use a retrieve-and-read pipeline, which first retrieves several passages from a corpus, and then applies reading comprehension models to extract (or generate) answers from the retrieved passages. However, passages found by retrieval models such as BM25 and DPR \citep{karpukhin2020dense} often lack diversity, i.e. the top-ranked passages commonly mention the same answers. While increasing the number of retrieved passages can increase recall of answers, it makes the answer prediction process more expensive---often quadratically more expensive.\footnote{The complexity of a Transformer-based reading comprehension model has the complexity of $O(N^2)$ where $N$ is the number of tokens.}

Rather than retrieving passages, an alternative way to implement a retrieve-and-read question-answering system is to extract information from a corpus, and retrieve the extracted information.  Several recent systems have proposed extracting information as question and answer (QA) pairs \cite{lewis-etal-2021-paq,chen2022qa,chen2022augmenting,wu2022efficient}. To answer open-domain questions, models retrieve generated QA pairs as evidence, rather than retrieving passages. Compared to passages, questions easier to retrieve, and are much shorter (usually 10 to 20 tokens, while passages  commonly contain hundreds of tokens). Therefore, more questions can be retrieved, to increase the recall of answers. 

\ifx\nodata\undefined
In this paper, we studied the use of generated QA pairs in answering ambiguous questions. First, we construct a new database of questions from Wikipedia, named \database{} (\dbexpansion{}), which contains 127 million questions and 137 million answers. 
Among the generated question, 5.8 million questions have more than one answer.\footnote{Some questions in \database{} are semantically the same. Those questions are not counted in the 5.8 million questions.} \database{} doubled the size of the previous database of question (PAQ), and more importantly, improves the recall of answers by 30\% on two open-domain QA datasets with ambiguous questions.\footnote{Please see \S \ref{sec:data_answer_coverage} for more information on the evaluation of answer coverage.} Second, we show that retrieving from \database{} leads to more diverse answers and therefore higher answer recall on two benchmark QA datasets with ambiguous questions, AmbigQA \citep{min2020ambigqa} and WebQuestionsSP \citep{yih-etal-2016-value}, with an improvement of up-to 6.7 point in recall@10. Third, we show that retrieving questions from \database{} improves the performance of the ASQA task \citep{stelmakh2022asqa}, a very challenging task of long-form question answering in summarizing and comparing answers of ambiguous questions. It improves the baselines by 1.9 points in DR and achieves a state-of-the-art on ASQA.

\else
In this paper, we study the use of QA databases in answering ambiguous questions. First, we describe methods for producing a large set of questions from Wikipedia.  In contrast to most previous question generation (QG) systems, our QG model is trained on questions from  AmbigQA \citep{min2020ambigqa}, which contains NQ \cite{kwiatkowski2019natural} questions that have been revised by adding additional information from passages to questions to reduce ambiguity.
This results in a very large set of questions, most of which have only one answer: 
we generate 127 million questions, with only 5.8 million that have more than one answer. Second, we show  this database of more-specific, better-grounded generated questions can be used to indirectly retrieve passages relevant to a question, by finding passages that generate questions similar to a query question $x$.  This retrieval approach leads to more diverse passages and higher answer recall on two benchmark QA datasets with ambiguous questions, AmbigQA \citep{min2020ambigqa} and WebQuestionsSP \citep{yih-etal-2016-value}. Third, we show that retrieving from generated questions, and incorporating these generated questions as context, leads to improvement on the ASQA task \citep{stelmakh2022asqa}, a very challenging task of long-form question answering for summarizing and comparing answers of ambiguous questions. Overall we improve the baselines by 1.9 points in DR and achieve a state-of-the-art on ASQA. 

\fi

\section{Related Work}

 In previous work, \citet{lewis-etal-2021-paq} constructed a database of 67M probably asked questions (PAQ) from Wikipedia and proposed to answer open-domain questions by retrieving similar questions from PAQ. Later work \citet{chen2022augmenting,wu2022efficient} proposed alternative approaches to using databases of QA pairs in QA systems, proposing pre-training methods that are more statistically-efficient \cite{chen2022augmenting} or more computationally efficient \cite{wu2022efficient}, and developing multi-hop QA models \cite{chen2022augmenting} that retrieve multiple times from a QA memory.   
 However, prior QA-memory based models \citep{lewis-etal-2021-paq,chen2022augmenting,wu2022efficient} focused on the traditional QA setting of unambiguous questions with unique answers.
 This leads to many differences in emphasis: for example, the PAQ dataset purposely removes generated questions with multiple answers.

Another efficient way to store world knowledge is to build knowledge bases (KB) or  knowledge graphs (KG), such as Freebase and Wikidata, where information is stored with entities and relations as triples, e.g. (``Charles Darwin'', ``author of'',  ``On the Origin of Species''). Knowledge bases have been commonly used in many knowledge intensive tasks due to its structured nature \citep{sun2018open,min2019knowledge}. Knowledge bases, however, lack the expressiveness in representing complex relationships that involve multiple pieces of information, and often do not contain information in a format that naturally reflects users' questions.


To resolve this problem, \citet{dhingra2020differentiable,sun2021reasoning} proposed to construct virtual knowledge bases that are not restricted to pre-defined vocabularies of entities and relations. \citet{dhingra2020differentiable} proposed to store entity-centric information as vectors and build a large database of vectors by iterating through passages in Wikipedia. \citet{sun2021reasoning} encoded pair-wise relationship between entities as vectors. Both methods support a similar reasoning process as regular knowledge bases.  
Others have argued for use of entity-linked QA pairs as a formalism for storing knowledge, as a representation that is more aligned with users' information needs, but still are closely related to traditional AI representations like KBs and KGs \cite{chen2022qa}.

Recent interest in QA for ambiguous questions poses new challenges for retrieving and representing knowledge. In such datasets, models are required to not only find one of the correct answers, but also comply with additional requirements associated with the need to choose between multiple answers. For example, Temp-LAMA \citet{dhingra2022time} requires models to answer time-sensitive questions under given time constraints; \citep{zhang2021situatedqa} contains questions with geographic and temporal constraints; ConditionalQA \citet{sun2022conditionalqa} contains constraints based on user scenarios; and ROMQA \cite{zhong2022romqa} requires QA subject to different combinations of constraints. 

This work builds especially on the AmbigQA dataset \cite{min2020ambigqa}, which contains NQ questions that have multiple interpretations, and the ASQA dataset \cite{stelmakh2022asqa}.  The ASQA dataset contains ambiguous questions with long-form answers, where ther answers explain in text what the alternative interpretations of the original question are, and what the answer is for each interpretation.


\ifx\nodata\undefined

\ifx\nodata\undefined
\section{\database}
We construct \database{} from Wikipedia which contains 127 million questions and 137 million answers. The construction process involves three stages: answer detection, question generation, and answer verification. We discuss each stage in detail in this section and compare each stage to another popular database of questions, PAQ \citep{lewis-etal-2021-paq}.
\else
\subsection{Question Generation from Wikipedia}
We generate questions and answers from Wikipedia passages. The construction process involves three stages: answer detection, question generation, and answer verification. We discuss each stage in detail in this section and compare each stage to another popular database of questions, PAQ \citep{lewis-etal-2021-paq}. We name our database of generated questions \database{} (\dbexpansion{}).
\fi

\ifx\nodata\undefined
\subsection{Source}
\else
\subsubsection{Source}
\fi
We use the dump of Wikipedia preprocessed by DPR \citep{karpukhin2020dense} as inputs to generate questions. In DPR's dump, Wikipedia articles are chunked into  passages which contain 100 tokens. The preprocessed dump of Wikipedia contains 21 million passages. Since many of the question interpretations in ASQA involve less popular entities, we generate questions from all 21 million passages. In contrast PAQ generates from only 9.1 million passages (filtered by a learned question selection model).

\ifx\nodata\undefined
\subsection{Stage 1: Answer Detection}
\else
\subsubsection{Stage 1: Answer Detection}
\fi
A Wikipedia passage usually contains multiple pieces of information and thus questions can be asked from different perspectives. We let the question generation process to be answer-conditioned to reflect this observation. During generation, possible answers are first detected and then questions are generated for every detected answer \citep{lewis-etal-2021-paq, chen2022augmenting}.

We model the answer detection step as a sequence-to-sequence (seq2seq) generation problem with a T5 model \citep{raffel2020exploring}.\footnote{We use the pretrained T5-11B model for all subtasks in constructing \database{} to ensure the high quality of data.} We do not use a Named Entity Recognition (NER) model for answer prediction, as used in PAQ \cite{lewis-etal-2021-paq}. The input of the generation task is a Wikipedia passage and the output is a text span that is likely to be the answer to some questions. The answer detection model (with a pretrained T5) is finetuned on NQ \citep{kwiatkowski2019natural}. 
We use beam search with beam size of 32 to generate multiple outputs, but filter these outputs with two heuristics. First, we require the generated spans to be sub-strings of the Wikipedia passage. Second, we merge spans which are identical after removing articles and punctuation. We end up with 283 million answers detected from 21 million Wikipedia passages.

\ifx\nodata\undefined
\subsection{Stage 2: Question Generation}
\else
\subsubsection{Stage 2: Question Generation}
\fi
Given answers detected from a Wikipedia passage, we then train a model to generate questions for the specific answers. Again, we finetune a T5 model for the question generation task. An input for question generation contains a passage and a target answer, e.g. ``\texttt{answer}: Michigan Stadium \texttt{context}: The Michigan Stadium is the home stadium ...''. An expected output should first repeat the target answer and then generate a question, e.g. ``\texttt{answer}: Michigan Stadium \texttt{question:} Where is the home stadium ...''. In preliminary experiments, encouraging the model to first repeat the answers generally improves the quality of questions by making generated questions more specific to target answers.

We use question and answer (QA) pairs from AmbigQA \citep{min2020ambigqa} to train the question generation task. Questions in AmbigQA originate from NQ but are revised by adding additional answer-specific information from passages to questions to remove ambiguity. This departs from most prior QG work, which generally trains models on SQuAD \citep{rajpurkar2016squad} or NQ \cite{kwiatkowski2019natural}. While AmbigQA is smaller than either of these datasets, the questions are more natural than SQuAD (where questions were \emph{formulated} by crowdworkers looking at the passage) and better-grounded than NQ (since questions are \emph{revised} by crowdworkers looking at the passaege), which seems to be a happy medium in producing natural questions with minimal hallucination\citep{bandyopadhyay2022improving}. 
\looseness=-1

We use greedy search at inference time to generate one question per answer. While PAQ used beam search (with a beam size of 4) to increase the number of generated questions \citep{lewis-etal-2021-paq}, we find that questions generated from beam search are often very similar to each other. Having near-duplicate questions makes the database larger but does not increase the utility of the database for most downstream tasks.


\ifx\nodata\undefined
\subsection{Stage 3: Answer Verification}
\else
\subsubsection{Stage 3: Answer Verification}
\fi
Questions generated from the previous step are sometimes invalid -- i.e. some questions may not be answerable from the provided passages, or the correct answers to the generated questions are different from the answers from which the questions are generated. Therefore, an additional answer verification step is needed.

We train a question answering (QA) model in the reading comprehension setting to perform the answer verification task. In particular, the model takes a passage and a generated question to predict an answer. If an answer does not exist in the passage, the model should predict ``not answerable''. We finetune a T5 model on SQuAD v2 \citep{rajpurkar2018know}, a reading comprehension dataset which contains unanswerable questions. During verification, we drop questions if their predicted answers are ``not answerable'' or different from their original answers.\footnote{We normalize the original and predicted answers before comparison using scripts provided by \citet{rajpurkar2016squad}.} After the verification step, 156 million questions are left.


The question generation process often produces questions that are ambiguous in an open-book setting, i.e. they have multiple answers.  This is expected since many NQ questions are themselves ambiguous \citep{min2020ambigqa} when considered carefully.  In PAQ, questions that have multiple open-book answers are filtered by running an open-book QA system and discarding questions with an open-book answer different from the one used for generation.  This has several disadvantages: it is expensive, since open-book QA is expensive to run; it is relatively noisier than our proposed QA-based filter, since open-book QA is less accurate than machine-reading style QA; and it filters out more than 76\% of the generated questions, and it is not actually appropriate for some downstream applications (such as the ones considered in \S \ref{sec:asqa_task}), where questions are used in conjunction with the passages from which they were generated.

\ifx\nodata\undefined
\subsection{Statistics}
\else
\subsubsection{Statistics}
\fi
We merge the question and answer pairs by merging pars with identical questions and end up with 127 million unique questions, among which 14.3 million questions have more than one answer mention, and 5.8 million questions have more than one unique answer.\footnote{Merging is performed by word matching, even though many questions are semantically same. }


\section{Coverage of Answers}
\label{sec:data_answer_coverage}

We measure the coverage of answers of \database{} on two benchmark datasets of ambiguous questions with multiple answers, AmbigQA \citep{min2020ambigqa} and WebQuestionsSP (WebQSP) \citep{yih-etal-2016-value}. We evaluate the exact match of answer spans after normalizing the answers. The recall of answers is shown in Table \ref{tab:stats_answer_recall}. The recall of \database{} is 6.2\% higher than PAQ on AmbigQA and 8\% higher on WebQuestionsSP (WebQSP). In addition, we measure the coverage of answers from NQ as a reference for non-ambiguous questions. \database{} improves the coverage of answers by 2.1\%.

\begin{table}[t]
\small
\centering
\begin{tabular}{lcccccc}
\toprule
          & \multicolumn{2}{c}{AmbigQA} & \multicolumn{2}{c}{WebQSP} & \multicolumn{2}{c}{NQ}  \\
          & Ans    & QA   & Ans   & QA   & Ans & QA \\ \midrule
PAQ       & 83.0      & 45.1                & 81.9     & 48.5   & 89.9 & 70.3         \\
\database & 89.2      & 79.3                & 89.9     & 82.5   & 92.0    & 85.6         \\
\bottomrule
\end{tabular}
\caption{Recall of answers of \database{} on AmbigQA (dev), WebQuestionsSP (test) and NQ (dev). ``Ans'' represents the recall of answers by matching answer strings (after normalization steps). ``QA'' is evaluated on 100 randomly selected question by human annotators by checking whether questions of the matched answers in the database are actually relevant. 
}
\label{tab:stats_answer_recall}
\end{table}

However, QA pairs in the databases that contain the answers are sometimes irrelevant to the questions asked. For example, (``What is the largest stadium in the US?'', ``the Michigan Stadium'') may not used to answer the question ``Where is the home stadium of the Michigan Wolverines?'' Therefore, the recall of answers is insufficient. To measure the true recall, we manually examine 100 questions to check whether the questions are relevant. We denote this metric as ``QA'' in Table \ref{tab:stats_answer_recall}. The true recall of QA pairs in \database{} is 30\% higher than PAQ on both AmbigQA and WebQuestionsSP (WebQSP) datasets. For non-ambiguous questions (NQ), the recall improves by 15.3\%.

\fi

\ifdefined\nodata
\section{Method}
In this section, we first discuss our approach to constructing a database of questions from Wikipedia, and then propose methods which use the generated questions for two important tasks in answering ambiguous questions: retrieving passages with diverse answers, and generating long-form answers to ambiguous questions.
\else
\section{\database{} for Ambiguous QA}
In this section, we propose methods which consume \database{} for two important tasks in answering ambiguous questions, including retrieving passages with diverse answers, and generating long outputs to discuss the difference between multiple answers for question disambiguation.
\fi


\ifdefined\nodata

\fi

\subsection{Retrieval of Diverse Passages} \label{sec:method_diverse_psg}
One common problem with existing retrieval models \citep{karpukhin2020dense,ni2021large} 
for open-domain QA is the lack of diversity of the retrieved results \cite{min2021joint}, i.e. only a subset of correct answers are obtained from the top-retrieved passages. This restricts models' performance in predicting multiple answers and in comparing different answers. We show that we can get more diverse passages \emph{indirectly}, by first retrieving similar generated questions $q'$ given a input question $x$, and then using as the final retrievals the passages from which the $q'$'s here generated. 

Retrieving questions $q'$ given a question $x$ is analogous to retrieving passages from text corpora,  soany existing retrieval method can be applied. In this paper, we use a sparse retrieval model, BM25, and a state-of-the-art dense retrieval model, GTR \citep{ni2021large}. GTR was originally designed for passage retrieval but the query encoder and passage encoder in GTR share parameters, so, we can directly use it to encode and retrieve questions as well.
We use GTR-large and finetune the checkpoint of GTR on NQ-open \citep{kwiatkowski2019natural} in our experiments.

Questions retrieved from \database{} are then mapped to passages where those questions were generated. With an input question $x$, the score for a passage $p_i$ is
\begin{equation}\label{eq:retrieval_max}
    s(x, p_i) = \textnormal{max}_{q'\in \textnormal{GEN}(p_i)}f(x, q')
\end{equation}
\noindent where $\textnormal{GEN}(p_i)$ is the set of questions generated from the passage $p_i$ and $f(x, q')$ is the retrieval score of the question $q' \in \textnormal{GEN}(p)$ from BM25 or GTR. We denote this method as ``max'' in the our experiments (\S \ref{sec:exp_diverse_psg}).


In addition, we propose another simple heuristic to map questions to passages. It returns passages from which the most top-$k$ retrieved questions are generated. We use $k = 50$ in our experiments. This method is denoted as ``count'' in our experiments (\S \ref{sec:exp_diverse_psg}).
\begin{equation}\label{eq:retrieval_count}
    s_c(x, p_i) = |\{\textnormal{GEN}(p_i) \cap \textnormal{argmax}_{k, q'}f(x, q')\}|
\end{equation}

\subsection{Ambiguous QA with Long Outputs}\label{sec:asqa_task}
\begin{table*}[t]
\small
\centering
\begin{tabular}{llll}
\toprule
Answers            & Context                                                                                             & Revision 1   & Revision2                                                                                                                       \\ \midrule
Michigan Stadium     & \begin{tabular}[c]{@{}l@{}}Michigan Stadium, nicknamed ``The\\ Big House'', is the home stadium for\\ the University of Michigan \textbf{\textit{men's}}\\ \textbf{\textit{football team}} (Michigan Wolverines)\\ in Ann Arbor, Michigan. Michigan \\Stadium was \textbf{\textit{built in 1927}}, and it is \\the largest stadium in the US...\end{tabular} & \begin{tabular}[c]{@{}l@{}}Where is the home stadium\\ of Michigan Wolverines \\ \textbf{\textit{men's football team}}? \end{tabular} & \begin{tabular}[c]{@{}l@{}}Where is the home stadium\\ of Michigan Wolverines \\ men's football team \\ \textbf{\textit{built in 1927}}? \end{tabular} \\ \midrule
Crisler Center  & \begin{tabular}[c]{@{}l@{}}Crisler Center (formerly known as the\\ University Events Building  and Crisler\\ Arena) is an \textbf{\textit{indoor}} arena located in\\ Ann Arbor, Michigan. It is the home\\ arena for the Michigan Wolverines \\ \textbf{\textit{men's and women's basketball teams}}...\end{tabular}               & \begin{tabular}[c]{@{}l@{}}Where is the \textbf{\textit{indoor}} home\\ stadium of Michigan Wolverines?\end{tabular}   & \begin{tabular}[c]{@{}l@{}}Where is the indoor home \\ stadium of Michigan Wolverines \\ \textbf{\textit{men's and women's basketball?}}\end{tabular}                  \\
\bottomrule
\end{tabular}

\caption{Examples of revisions of a question ``Where is the home stadium of Michigan Wolverines?'' (not an exclusive list). Depending on different answers, questions are revised at each revision step to add additional answer-specific information. We consider question revision as a question expansion step which moves information from passages to questions.}
\label{tab:data_example}
\end{table*}

In the second task, we investigate the challenging task of answering ambiguous questions with long outputs summarizing multiple answers to the questions.
For ambiguous questions that have different answers, one practical way to answer such questions is to specify under what conditions answers are correct. For example, for the question ``Where is the home stadium of Michigan Wolverines?'', in addition to predicting a list of answers, \{``Crisler Center'', ``Michigan Stadium'', ...\}, a QA system should clarify that ``Crisler Center'' is the home stadium of the Michigan basketball team while the ``Michigan Stadium'' is the home of the football team. The ASQA task proposed to answer ambiguous questions by summarizing the multiple answers into short paragraphs, e.g. ``The home stadium of Michigan Wolverines men's football is the Michigan Stadium, while the stadium of its men's basketball team is the Crisler Center. Crisler Center is also the home stadium for Michigan Wolverines women's basketball''.

Previous models simply retrieve passages from a text corpus and generate answers from the retrieved results. However, the retrieved passages are usually long and contain information irrelevant to the answers.
We propose to retrieve questions from \database{} as a concise representation of question-specific information from passages. We additionally propose a question revision step which operates on the retrieved questions to include more detailed information for the disambiguation task.



\subsubsection{Question Revision}\label{sec:cond_gen}


While the questions in \database{} are fairly unambiguous, we also explored approaches to make the questions include more information from the passages from which they were generated.
We trained a sequence-to-sequence model to extract answer-specific information from passages where \database{} questions are generated and rewrite the questions to include such information. Examples of questions before and after revision are shown in Table \ref{tab:data_example}: e.g., the model locates the information ``men's football'' from the context ``... is the home stadium for the University of Michigan men’s football team (Michigan Wolverines) in Ann Arbor ...'' and adds it to the initial question. The revised question, ``Where is the home stadium of the Michigan Wolverines men's football team built in 1927?'', contains information \{``men's football'', ``built in 1927''\} that is specific to the answer ``Michigan Stadium''. Compared to passages with hundreds of tokens, the revised questions are more concise in capturing information that is specific to answers. 

We finetune a T5 model to perform the question revision task. The T5 model is trained with data provided as auxiliary information in the ASQA dataset \cite{stelmakh2022asqa}, which contains revised questions for different answers $a_i$ and passages $p_i$ provided to human annotators to write the revised questions $q'_i$.\footnote{The revised questions originate from the AmbigQA \cite{min2020ambigqa} dataset. ASQA conducted additional annotation and included more auxiliary information, such as passages for question revision.} The question revision model takes an ambiguous question $q$, an answer $a_i$, and a passage $p_i$ to generate a revised question $q'_i$. The input and output of the model are shown below.
\begin{gather*}
    \textnormal{input} = ~\textnormal{question:}~ q + \textnormal{answer:}~ a_i + \textnormal{passage:}~ p_i \\
    \textnormal{output} =  ~\textnormal{answer:} ~ a_i + \textnormal{revised:}~ q'_i
\end{gather*}


At inference time, we repeat the revision process $k$ times to increase the amount of information added to original questions. In the experiments, we use $k=2$ because we observe the model tends to generate identical questions if $k > 2$. The revised questions have an average length of 14.5 compared to original questions, which average 9.0 words long.

\subsubsection{Long-form Answer Generation}
After revision of the top-retrieved \database{} questions, we perform a generation task, to summarize the differences between multiple answers of the ambiguous questions.
In addition to the revised questions from \database{}, we also retrieve a few passages from Wikipedia for generating long-form answers. We find retrieving passages is necessary for ASQA---perhaps because during annotation annotators were encouraged to include background information in the long-form answers. Such information is not specific to any answer, so merely retrieving from \database{} does not provide the necessary information.
To mitigate this problem, we follow the baseline to also include top $n$ passages retrieved by JPR \cite{min2021joint} from Wikipedia \cite{stelmakh2022asqa}.\footnote{JPR is an auto-regressive reranking model aiming for increasing the diversity of retrieved passages.} The inputs to the generation model are thus a concatenation of the original question $q$, answers and retrieved questions $\{(a_i, q'_i)\}$,  and retrieved passages $\{p_j\}$. The target outputs are the long answers provided in ASQA. We finetune a T5-large model \cite{raffel2020exploring} for this generation task.
\begin{align*}
    \textnormal{input} = ~&\textnormal{question: $q$ + conditions: $a_1$, $q'_1$, ...}
    \textnormal{ + passages: $p_1$, ...}
\end{align*}


\section{Experiments}
In this section, we discuss the experimental results for retrieving diverse passages and generating long-form answers for ambiguous questions.

\subsection{Retrieval of Diverse Passages} \label{sec:exp_diverse_psg}
\subsubsection{Dataset}
We use AmbigQA \cite{min2020ambigqa} and WebQuestionsSP \cite{yih-etal-2016-value} in our experiments. AmbigQA is an open-domain QA dataset derived from NQ \cite{kwiatkowski2019natural} which contains questions that are ambiguous and thus have multiple possible answers. WebQuestionsSP (WebQSP) \citep{yih-etal-2016-value} is another dataset which contains open-domain questions asked by web users, and a subset of the questions have multiple answers. We only evaluate on multi-answer questions (in both datasets) in this experiment: 1172 questions in the AmbigQA dev set and 809 questions in the WebQuestionsSP test set have multiple answers.\footnote{We consider questions have multiple answers if at least one of the annotators find multiple answers.}

\subsubsection{Evaluation} To measure the diversity of retrieval, we evaluate models' performance as the recall of answers. Similar to traditional passage-level retrieval models \citep{karpukhin2020dense}, the recall is measured as the percentage of correct answers that are mentioned in the retrieved passages. 

\begin{table}[t]
\centering
\small
\begin{tabular}{lHHcccc}
\toprule
& \multicolumn{4}{c}{AmbigQA}  & \multicolumn{2}{c}{WebQSP} \\
k             & 1    & 3    & 5    & 10   & 5   & 10  \\ 
\midrule
\textit{passage - based retrieval} & & & & & & \\
Wikipedia + BM25          & 18.3   & 29.4    & 35.5    & 44.4    & 23.4 & 32.0 \\
Wikipedia + DPR          & 29.3 & 44.7 & 50.7 & 57.7 & 36.2  & 43.0  \\
Wikipedia + GTR          & 34.8 & 49.2 & 55.0 & 61.9 & 38.8  & 46.3  \\
\midrule
\textit{question - based retrieval} & & & &   &   &   \\
PAQ + BM25 (max)   & 23.0 & 31.8 & 36.9 & 43.6 & 26.7  & 32.7  \\
PAQ + GTR (max)   & & & 43.7 & 51.3 & 33.2  & 39.6  \\
\database{} + BM25 (max)   &      &      & 35.5     & 45.8     & 24.8  & 34.4  \\
\database{} + GTR (max)    & 31.1     &      & 53.6 & 60.4     & 45.3  & 51.8  \\
\database{} + BM25 (count) &      &      & 36.4     & 47.0     & 25.7  & 35.8  \\
\database{} + GTR (count)  & 32.5     &      & \textbf{55.9} & \textbf{63.4}     & \textbf{46.7}  & \textbf{53.0}  \\
\bottomrule
\end{tabular}
\caption{Recall@k of retrieving diverse answers for multi-answer questions in AmbigQA and WebQuestionsSP. Experiments with ``max'' (Eq. \ref{eq:retrieval_max}) and ``count'' (Eq. \ref{eq:retrieval_count}) use different methods to map top-retrieved questions to passages. We use GTR-large in the baselines and our method.}
\label{tab:ambigqa_retrieval}
\end{table}

\subsubsection{Results}
Experimental results are presented in Table \ref{tab:ambigqa_retrieval}. Numbers in the first block (passage-based retrieval) show the performance of baseline models, BM25, DPR \citep{karpukhin2020dense} and GTR \citep{ni2021large}, in directly retrieving passages from Wikipedia. DPR is another popular dense retrieval method with separate query and candidate encoders, and trained with hard negatives. We re-run the DPR open-sourced code and evaluate the retrieved results. We also run GTR-large for passage retrieval on both datasets. For question-based retrieval, we apply the proposed method on both PAQ \citep{lewis-etal-2021-paq} and our  \database{} dataset. Again, we use GTR-large in our method. Experiments with (max) refer to the method of directly mapping top-retrieved questions to passages where they are generated (Eq. \ref{eq:retrieval_max}), while ones with (count) refer to returning passages where most top-retrieved questions are generated (Eq. \ref{eq:retrieval_count}). Compared to passage-based retrieval methods, indirect retrieval with \database{} yields better performance than using BM25 or GTR. In particular, the recall@10 with BM25 improves from 44.4 to 45.8 on AmbigQA and from 32.0 to 34.4 on WebQuestionsSP. The performance with GTR is also better with \database{}. On AmbigQA, the recall@10 improves 61.9 to 63.4. More improvement comes on WebQuestionsSP with an increase from 46.3 to 53.0. We conjecture that the improvement with GTR is less significant on AmbigQA because GTR is pretrained on NQ, which is a superset of AmbigQA.

\subsection{Ambiguous QA with Long Outputs}

\begin{table*}
\centering
\small
\begin{tabular}{lccccc}
\toprule
                           & LEN (words) & ROUGE-L & STR-EM & DISAMBIG-F1 & DR    \\ \midrule
DPR @ 1$^\dagger$                    & 99.9        & 31.1    & 30.1   & 16.7        & 22.8  \\
JPR @ 1$^\dagger$                    & 196.8       & 27.9    & 45.0   & 25.8        & 26.9  \\ \midrule
T5 (1 passage)$^\dagger$             & \textbf{63.0}        & 40.3   & 33.6   & 21.2        & 29.2 \\
T5 (3 passages)$^\dagger$             & 71.1        & 42.7  & 39.9   & 25.1        & 32.7 \\
T5 (5 passages)$^\dagger$             & 71.6        & 43.0   & 41.0   & 26.4        & 33.7 \\
T5 (5 passages) *           & 68.1        & 43.0    & 40.1   & 26.4        & 33.7      \\
T5 (7 passages) *           & 69.3        & 43.0    & 39.5   & 25.5        & 33.1      \\
T5 (10 passages) *          & 68.9        & 43.0    & 39.2   & 25.9        & 33.2      \\ \midrule
\textit{ours} & & & & & \\
T5 (1 passage + 10 questions) & 58.3        & 41.6    & 39.4   & 26.5        & 33.2      \\
T5 (2 passages + 10 questions) & 62.0        & 42.9    & 41.8   & 28.0        & 34.6      \\
T5 (3 passages + 10 questions) & 63.3        & 42.9    & 41.5   & 28.2        & 34.8      \\
T5 (5 passages + 10 questions) & 63.5        & \textbf{43.8}    & \textbf{42.4}   & \textbf{28.9}        & \textbf{35.6}     \\ \midrule
T5 (oracle)                & 82.6        & 46.6    & 88.7   & 59.2        & 52.5 \\
Human                      & 64.8        & 49.4    & 98.4   & 77.4        & 61.8 \\
\bottomrule
\end{tabular}
\caption{Performance of long-form answer generation with retrieved answers and passages. All models are finetuned on T5-large. $^\dagger$ Numbers copied from the baseline \cite{stelmakh2022asqa}. * Numbers obtained by re-implementing the baseline models.}
\label{tab:asqa_main}
\end{table*}

\subsubsection{Dataset}
The ASQA dataset contains 4353 train and 948 dev examples. Each example contains an ambiguous question, a list of disambiguated questions and answers (short text spans), and a long-form answer which discusses the difference between short answers. Due to the high variance of long-form answers, each example in ASQA was annotated by two human annotators and the better score among the two annotations is recorded. The average length of answers is 65.0 white-space split tokens. Each question has an average of 3.4 different short answers.

\subsubsection{Evaluation}
In ASQA, predicted outputs are evaluated from a few different perspectives. First, as a long output prediction task, it evaluates the similarity of predicted outputs with reference outputs with ROUGE-L scores. Second, it measures the recall of answers in the predicted outputs (named STR-EM)--all possible answers must be mentioned in the predicted output in order to receive full STR-EM scores. Third, it introduces a learned metric DISAMBIG-F1, with the goal of measuring whether the disambiguating information about answers in the outputs is accurate. To compute DISAMBIG-F1, the ASQA dataset uses a learned a QA model to find the answers of a sequence of disambiguated questions (provided by annotators) from the generate output. The output will receive a full DISAMBIG-F1 score if all predicted answers from the QA model match the oracle answers of the disambiguated questions. Finally, they compute an overall score, DR, as the geometric mean of ROUGE-L and DISAMBIG-F1. In addition, LEN (words) measures the average length of outputs in terms of words. Shorter outputs with higher DR scores are preferred.

\subsubsection{Results}

We evaluate the finetuned T5-large model on the quality of predicted long-form answers. To show the effectiveness of the retrieved questions and answers from \database{}, we compare to the outputs generated from retrieved passages only. 

Results are presented in Table \ref{tab:asqa_main}. The first group of results (DPR@1 and JPR@1) means we directly return the top 1 passage retrieved by DPR and JPR. The second group of results, e.g. T5 (5 passages), shows the performance of directly generating outputs with the top 5 retrieved passages with T5-large. Both groups of numbers are copied from the original paper by \citep{stelmakh2022asqa}.

To check whether higher recall from more retrieved passages leads to better outputs, we re-implement the baseline model to run it on more retrieved passages. The results with 5 passages from our implementation matches the numbers reported in the original paper \cite{stelmakh2022asqa}. However, as shown in Table \ref{tab:asqa_main}, as the number of passages increases, both STR-EM and DISAMBIG-F1 drops (40.1 to 39.2, 26.4 to 25.9). 

In the third group of experiments, we retrieve questions from \database{} and add top 10 answers with their conditions to the input. Without changing the model, the performance of long-answer generation with information retrieved from \database{} increases by 0.8 in ROUGE-L and 2.5 in DISAMBIG-F1. In addition, the output length of the model with information from \database{} is also $\sim$10\% shorter than the baseline but covers more answers in its outputs.

\subsubsection{Ablations} 

\begin{figure}
\centering
\includegraphics[width=0.9\linewidth]{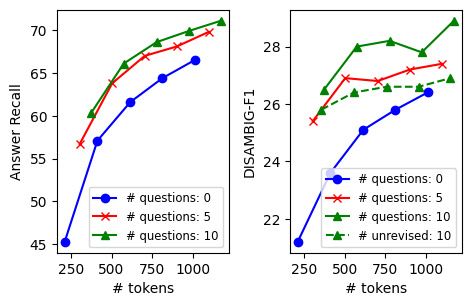}
\caption{\small\label{fig:asqa_abl} Left: Recall of answers from the retrieved results with varying number of passages.  Right: DISAMBIG-F1 of predicted long answers with varying number passages. Both figures show that inputs with more questions yield better outputs under certain number of tokens, in both answer recall and DISAMBIG-F1.}
\end{figure}

\textbf{Recall of Answers}
In the first ablation experiment, we justify the claim that retrieving questions from \database{} can improve the diversity of answers. 
We report the recall of answers with and without questions retrieved from \database{} in terms of numbers of tokens (see Figure \ref{fig:asqa_abl}). 
With 10 questions from \database{} added to 5 passages, the recall of answers improve from 65.5 to 71.1, which leads to around 2 additional points in the final DR metric.
From another perspective, with as few as 10 questions and 2 passages, the recall becomes comparable to 5 passages (66.1 vs. 66.5). Furthermore, the total length of 10 answers plus 2 passages is 43\% less than 5 passages (574.5 vs. 1008.8), since information from the revised questions are more concise. This eventually leads to 1 point of increase in the final DR metric (34.6 vs. 33.7) as shown in Table \ref{tab:asqa_main}. 

\textbf{Accuracy vs. Input Length}
We additionally compare the performance of models in DISAMBIG-F1 under different numbers of tokens. Results are shown in Figure \ref{fig:asqa_abl} (right). With \database{} questions, models get better DISAMBIG-F1 performance with shorter inputs. The DISAMBIG-F1 with 10 questions and 3 passages (775.6 tokens) is 28.2, better than the DISAMBIG-F1 of 27.2 with 5 questions and 4 passages (899.9 tokens) and DISAMBIG-F1 of 26.4 with 0 question and 5 passages (1008.8 tokens).

\textbf{Revised vs. Unrevised Questions} We further ablate our model to investigate the importance of question revision step proposed in \S \ref{sec:cond_gen}. The results are shown in Table \ref{fig:asqa_abl} (right). The model's performance with 10 revised question is consistently better than with unrevised questions. 

\textbf{Question Disambiguation}
In the next ablation experiment, we experiment with a more straightforward task to investigate whether the additional information retrieved from \database{} help disambiguating questions. Here, we study a question revision sub-task, using auxiliary data provided in ASQA.\footnote{This information is also available in the question revision sub-task in AmbigQA \citep{min2020ambigqa}.} 
In this sub-task, the model should revise an ambiguous question using information provided a passage such that it can differentiate the provided answer with others.
The task is similar to revising questions when constructing \database{} (\S \ref{sec:cond_gen}), except that the additional information added to the revised question should be contrastive, i.e. it should differentiate its answer with others possible answers to the ambiguous questions. For example, to differentiate the answer ``the Michigan Stadium'' and ``Crisler Center'', one should provide the additional information ``\textit{basketball team}'' vs. ``\textit{football team}'', but not ``\textit{built in 1927}'' vs. ``\textit{basketball team}''. 

A naive model simply takes an ambiguous question, an answer, and the provided passage as input to predict the revised question. We instead retrieve similar questions, along with their answers and conditions from \database{}, and augment the provided passage with the retrieved information, similar to \S \ref{sec:asqa_task}. The additional information should provide background knowledge for models to determine how to revise the question. Again, we finetune a T5 model for this question revision task. The model's output is measured by a metric, ``EDIT-F1'', proposed by \citet{min2020ambigqa}, to only evaluate the edits made in the revised questions.\footnote{Please refer to \citet{min2020ambigqa} for more information on ``EDIT-F1''.} 

The results are shown in Table \ref{tab:ambigqa_main}. In addition to the naive baseline of only taking the provided passage as input (passage-only), we experiment with a few other options that can potentially improve the coverage of information for the question revision task. First, we retrieve the top 1 passage from Wikipedia (top-1 Wikipedia). Second, we expand the provided passage with preceding and following tokens to double the length of inputs (adjacent context). Third, we deliberately add a passage which contains different answers of the same question (contrasting passage).\footnote{Also available as auxiliary information in ASQA.} Results in Table \ref{tab:ambigqa_main} shows that adding questions retrieved from \database{} is the most effective method in revising questions, and therefore justify our claim that questions from \database{} are concise and can provide sufficient background information for models to differentiate answers. 

\begin{table}[t]
\centering
\small
\begin{tabular}{lc}
\toprule
                 & EDIT-F1 \\ \midrule
oracle passage     & 22.4    \\
 + adjacent context   & 22.6        \\
 + top-1 Wikipedia      & 21.6        \\
 + contrasting passage & 23.0        \\ \midrule
 + top-5 \database{} questions             & 24.8    \\
 + top-10 \database{} questions             & \textbf{25.6}    \\
\bottomrule
\end{tabular}
\caption{Results on the question revision task of AmbigQA. We compare different approaches to increase the coverage of information in the inputs.}
\label{tab:ambigqa_main}
\end{table}

\section{Conclusion}
\ifx\nodata\undefined
In this paper, we proposed \database{}, a new database of 127 million questions constructed from Wikipedia, as an efficient method to store knowledge. We experiment \database{} on two different tasks to show its efficacy in solving questions with multiple answers. In the first task, we show that retrieving from \database{} can increase the diversity of retrieval results as an increase in the recall of answers. In the second task, we show that the increase in recall, along with the concise information contained in the \database{} questions, improves the performance of models on a challenging long-form QA task in summarizing and comparing different answers of ambiguous questions. In conclusion, we claim that \database{} is a powerful database. We expect people can use it on more tasks in future research.
\else
In this paper, we proposed to use a database of questions constructed from Wikipedia to answer ambiguous questions. We experiment on two different tasks to show its efficacy in solving questions with multiple answers. In the first task, we show that retrieving from generated questions can increase the diversity of retrieval results as an increase in the recall of answers. In the second task, we show that the increase in recall, along with the concise information contained in the revised questions, improves the performance of models on a challenging long-form QA task in summarizing and comparing different answers of ambiguous questions. 
\fi





\end{document}